\documentclass[10pt,twocolumn,letterpaper]{article}

\usepackage{cvpr}
\usepackage{times}
\usepackage{epsfig}
\usepackage{subfig}
\usepackage{graphicx}
\usepackage{amsmath}
\usepackage{amssymb}
\usepackage{tabularx}
\usepackage{caption}
\newcolumntype{P}[1]{>{\centering\arraybackslash}p{#1}}

\usepackage[breaklinks=true,bookmarks=false,colorlinks=true,citecolor=green,linkcolor=red]{hyperref}

\cvprfinalcopy 

\def\cvprPaperID{****} 

\ifcvprfinal\fi
\begin{document}

\title{NoPeopleAllowed: The Three-Step Approach to Weakly Supervised Semantic Segmentation}

\author{
Mariia Dobko $^{1, 2}$  \\
{\tt\small dobko\_m@ucu.edu.ua} \\
\and Ostap Viniavskyi $^1$  \\
{\tt\small viniavskyi@ucu.edu.ua}
\and Oles Dobosevych $^1$  \\
{\tt\small dobosevych@ucu.edu.ua}
\and\vspace*{-3ex}
{$^1$ The Machine Learning Lab, Ukrainian Catholic University, Lviv, Ukraine}
\and
$^2$ SoftServe, Lviv, Ukraine}



\thispagestyle{empty}

\twocolumn[{%
\renewcommand\twocolumn[1][]{#1}%
\maketitle
\begin{center}
    \vspace*{-0.5cm}
    \centering
    \includegraphics[width=\textwidth,height=2.7cm]{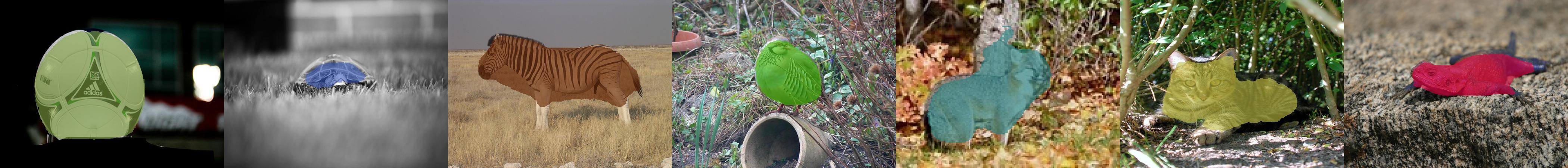}
    \captionof{figure}{The results of the proposed approach on the unseen data.}
\end{center}

\label{fig:examples}}]

\begin{abstract}
   We propose a novel approach to weakly supervised semantic segmentation, which consists of three consecutive steps. The first two steps extract high-quality pseudo masks from image-level annotated data, which are then used to train a segmentation model on the third step. The presented approach also addresses two problems in the data: class imbalance and missing labels. Using only image-level annotations as supervision, our method is capable of segmenting various classes and complex objects. It achieves 37.34 mean IoU on the test set, placing 3rd at the LID Challenge in the task of weakly supervised semantic segmentation.
   
\end{abstract}


\section{Introduction}

Deep learning methods have proven their efficiency in a variety of computer vision tasks, including semantic segmentation. However, their application to semantic segmantation typically requires large amounts of data with pixel-level annotations. We can, however, overcome this issue by developing weakly supervised methods that rely only on image-level labels. Omitting usage of pixel-wise annotations also provides considerable advantages. For example, it is less expensive and less time consuming; Lin et al. \cite{lin2014microsoft} calculated that collecting bounding boxes for each class is about 15 times faster than producing a ground-truth pixel-wise segmentation mask; collecting image-level labels is even more time-efficient. Working with image-level annotations also decreases the probability of disagreement between experts, since pixel-wise annotations tend to be less accurate and have a higher variance among labelers. 

\textbf{Data:} The dataset used was proposed on the LID Challenge \cite{lid2020}. It consists of 456,567 training images of objects from 201 classes including background. The validation and test sets have pixel-wise annotations, which are publicly available only for a validation set. The images in the train set are provided exclusively with image-level annotations. Moreover, the data have a lot of missing labels, and are also highly imbalanced towards three classes: `dog', `bird', and `person'. 

The class `person' has a large impact on other classes in the data; it usually appears in combination with others and often overlaps with such classes as `microphone', `sunglasses', `unicycle' etc. It is thus crucial to have correct labels for the class 'person'; however, the opposite is observed in the data: the image-level labels for this class are often missing, creating an additional challenge for the task. So, not only are the data biased towards a certain class, but they also suffer from imperfect labelling. These problems are usually present in many datasets, so a solution overcoming them will make an essential contribution to a larger field. 

We propose a novel weakly-supervised approach to semantic segmentation that uses only image-level annotations and deals with data that have severe class imbalance. It scores 37.34 mean Intersection over Union (IoU) on the test set placing third in the LID Challenge.

\section{Related Work}

In this paper, we follow the self-supervised paradigm of weakly supervised semantic segmentation, which suggests training a fully supervised segmentation model on the pseudo-labels generated from a classifier network. The image-level annotations are used to train a classifier; Class Activation Maps (CAM) \cite{zhou2016learning} are extracted afterward. Assessment of quantitative performance on PASCAL VOC 2012 \cite{pascal-voc-2012} validation set shows that the top five methods of weakly-supervised segmentation use the self-supervised learning approach \cite{chan2019comprehensive}. The nature of PASCAL VOC 2012 \cite{pascal-voc-2012} dataset is similar to the LID Challenge data, thus, we use the lessons learned on PASCAL VOC 2012 \cite{pascal-voc-2012} when developing a solution for the challenge.   
\par
Many methods of self-supervised learning for weakly supervised semantic segmentation have been recently suggested. Kolesnikov et al. \cite{kolesnikov2016seed} propose Seed Expand Constrain (SEC) method, which trains a Convolutional Neural Net (CNN), applies CAM to produce pseudo-ground-truth segments, and then trains a Fully Convolutional Network (FCN) optimizing three losses: one for the generated seeds, another for the image-level label, and, finally, a constraint loss against the maps processed by Conditional Random Fields (CRF). Huang et al. \cite{huang2018weakly} introduce Deep Seeded Region Growing (DSRG), which propagates class activations from high-confidence regions to adjacent regions with a similar visual appearance by applying a region-growing algorithm on the generated CAM. Another approach, proposed by Ahn et al. \cite{ahn2019weakly}, suggests using Inter-pixel Relation Network (IRNet) \cite{ahn2019weakly}, which takes the random walk from low-displacement field centroids in the CAM up until the class boundaries as the pseudo-ground-truths for training an FCN. Ahn et al. \cite{ahn2019weakly} focus on the segmentation of the individual instances estimating two types of features in addition to CAM: a class-agnostic instance map and pairwise semantic affinities. We incorporate IRNet \cite{ahn2019weakly} into one of the steps of our approach.

\section{Method}
The proposed approach consists of three consecutive steps: Classification followed by CAM generation, IRNet \cite{ahn2019weakly} for activation map improvement, and Segmentation. Each of these steps is followed by post-processing and improves the results of the previous one. All the experiments were performed on three Nvidia GeForce RTX 2080 TI GPUs. 

\subsection{Classification}
On the first step, we train fully-supervised classification models with image-level labels.

\textbf{Input:} 
We remove the 'person' class labels and balanced the other 199 classes (without background) using the downsampling technique. The obtained data is split into train and validation parts with 72,946 and 12,873 samples in each.

\textbf{Neural network architecture and loss}: For this step, we choose VGG16 arhitecture with 4 additional convolutional layers at the end, as proposed by Jiang et al. \cite{jiang2019integral}. We use binary cross-entropy loss for each output.

\textbf{Training procedure:} The model is trained with Adam optimizer \cite{kingma2014adam} and the learning rate $10^{-4}$ for the pretrained part and $10^{-3}$ for 4 extra convolutions. The input images are augmented using strong augmentation (horizontal flip, shift, scale, rotate, Gauss noise, random brightness and contrast, median blur, RGB shift).

\subsection{IRNet} 
For the second step, we choose IRNet \cite{ahn2019weakly} with Class Boundary Map and Displacement Field branches. The IRNet allows to improve boundaries between different object classes. It is trained on the generated maps from the first step and does not require extra supervision. This step allows us to obtain better pseudo-labels before proceeding to segmentation.

\textbf{Input:} As an input for IRNet  \cite{ahn2019weakly}, we chose only images from the train dataset that had confidence score of classification more than 0.8 and increase their amount by including the scaling with factors 0.5, 1, 1.5, and 2. All CAM are also postprocessed with CRF.

\textbf{Neural network architecture and loss:} As in the original paper we use ResNet50 \cite{he2016deep} concatenated activations from different layers as an architecture and the sum of Affinity loss and Displacement loss for a loss function.

\textbf{Training procedure:} The model is trained with freezed backbone and Stochastic gradient descent (SGD) optimizer with learning rate 0.05 for Displacement field part and learning rate 0.005 for Class Boundary Map. The same strong data augmentation as in Classification step is used.

\subsection{Segmentation} 
The segmentation step is done in a classic manner with masks obtained on a previous step.

\textbf{Neural network architecture and loss:} We use DeepLabv3+ \cite{chen2018encoder} with ResNet50 \cite{he2016deep} encoder that was pretrained on ImageNet and has stride replaced with dilation to increase receptive field. For output we use binary categorical cross-entropy loss.

\textbf{Training procedure:} The model is trained with SGD optimizer with learning rate 0.001, momentum 0.9 and weight decay $10^{-6}$. 

\subsection{Postprocessing}
The final prediction is made by averaging the predictions after horizontal flip and scale (factors: 0.5, 1, and 2). We refer to this technique as Test Time Augmentation (TTA).

\begin{figure*}[ht!]
    \centering
    \subfloat {
    \includegraphics[width=0.17\linewidth]{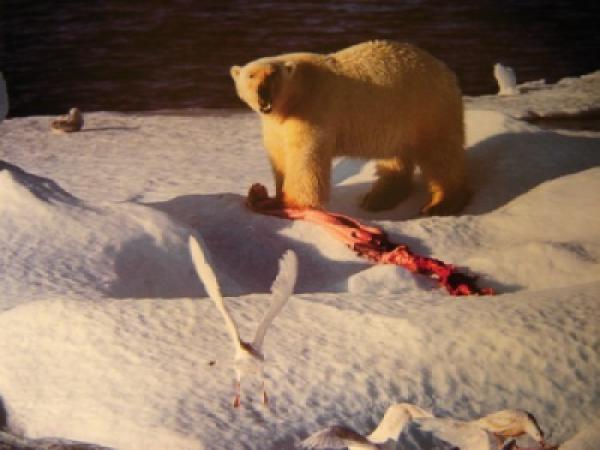}
    }
  \subfloat {
    \includegraphics[width=0.17\linewidth]{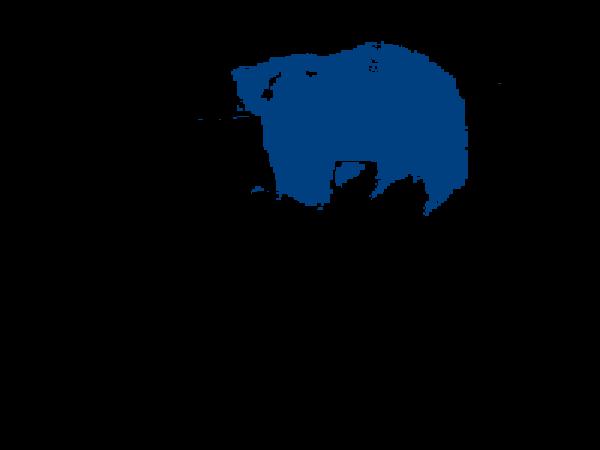}
    }
  \subfloat {
    \includegraphics[width=0.17\linewidth]{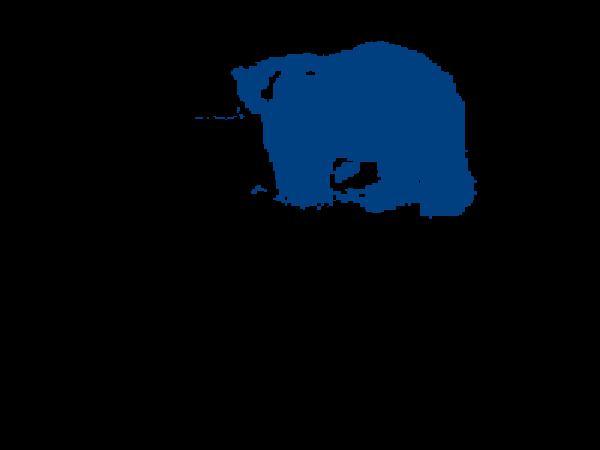}
    }
  \subfloat {
    \includegraphics[width=0.17\linewidth]{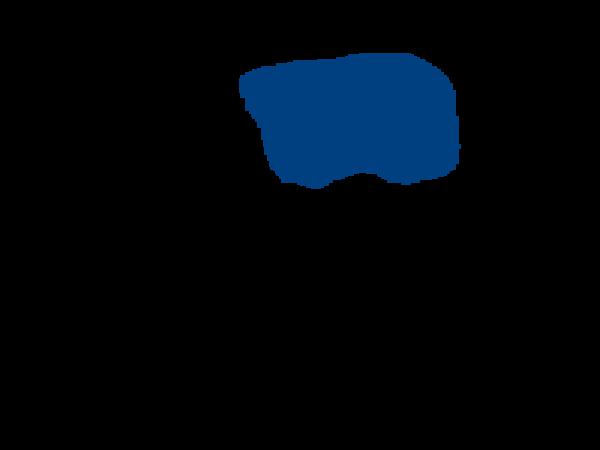}
    }
 \subfloat {
    \includegraphics[width=0.17\linewidth]{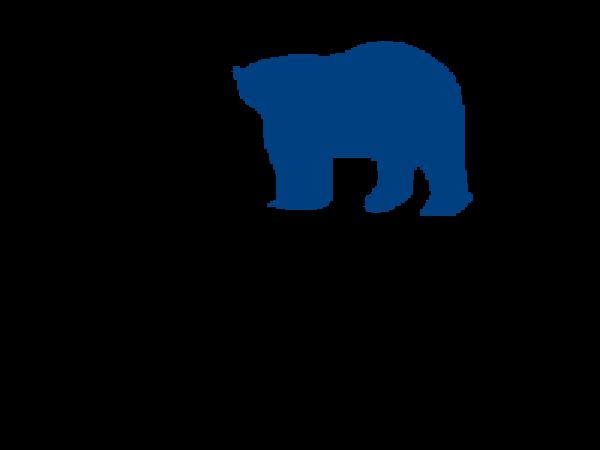}
    }
     \\[-1.5ex]
 \subfloat {
    \includegraphics[width=0.17\linewidth]{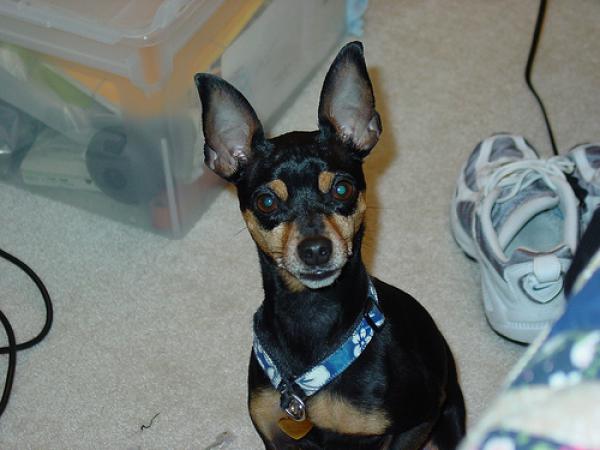}
    }
  \subfloat {
    \includegraphics[width=0.17\linewidth]{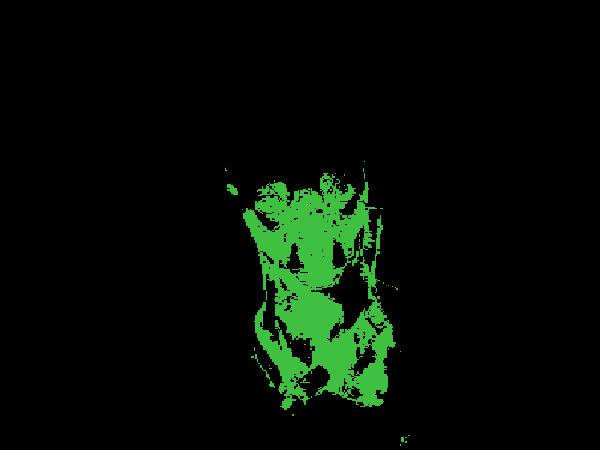}
    }
  \subfloat {
    \includegraphics[width=0.17\linewidth]{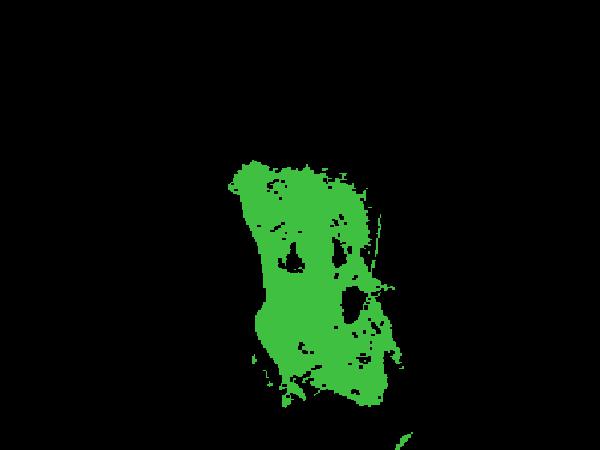}
    }
  \subfloat {
    \includegraphics[width=0.17\linewidth]{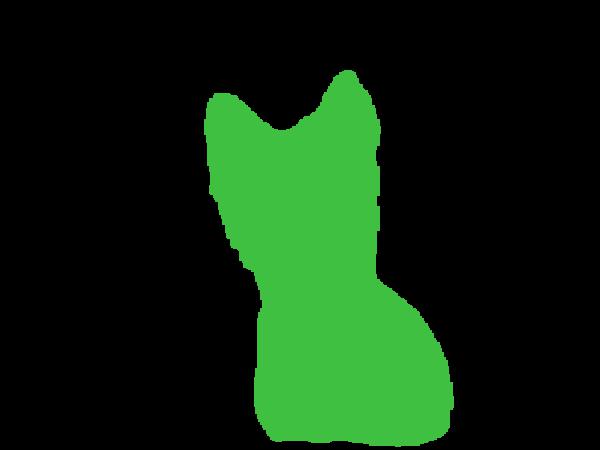}
    }
 \subfloat {
    \includegraphics[width=0.17\linewidth]{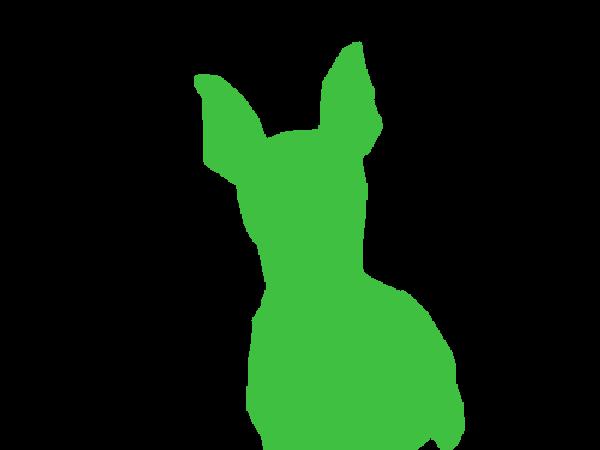}
    }
         \\[-1.5ex]
 \subfloat {
    \includegraphics[width=0.17\linewidth]{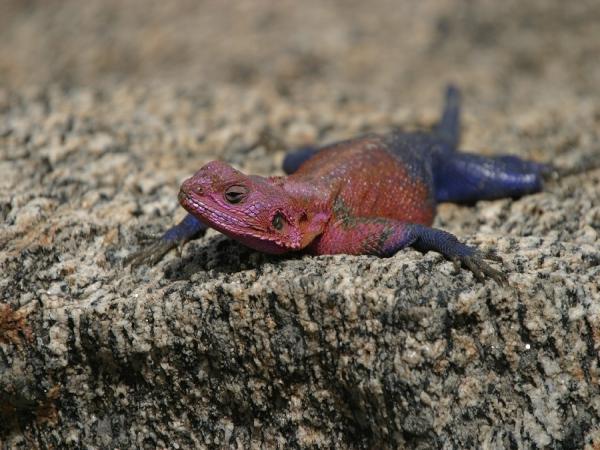}
    }
  \subfloat {
    \includegraphics[width=0.17\linewidth]{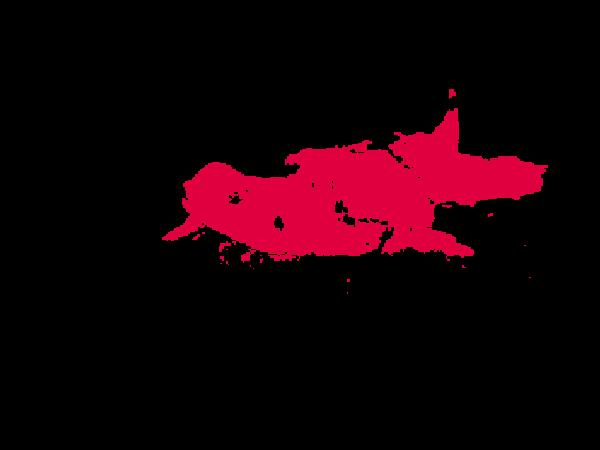}
    }
  \subfloat {
    \includegraphics[width=0.17\linewidth]{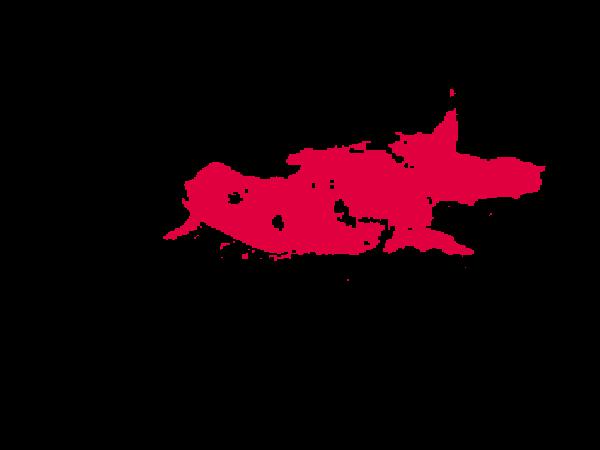}
    }
  \subfloat {
    \includegraphics[width=0.17\linewidth]{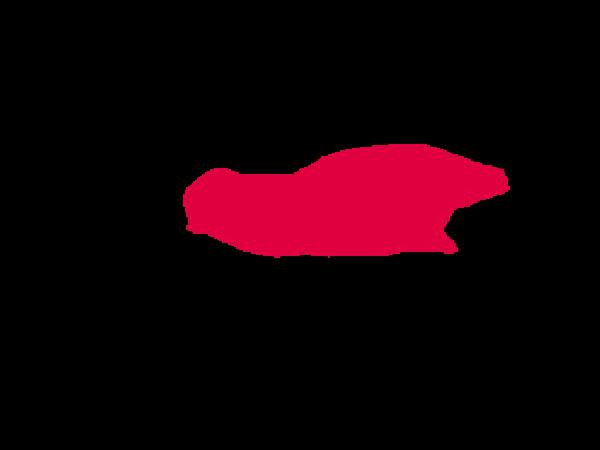}
    }
 \subfloat {
    \includegraphics[width=0.17\linewidth]{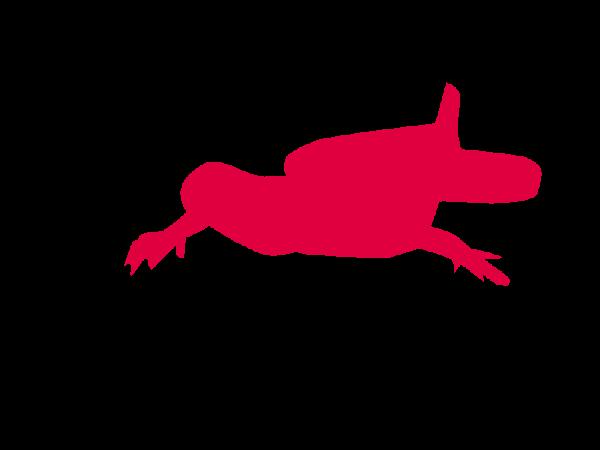}
    }
         \\[-1.5ex]
 \subfloat {
    \includegraphics[width=0.17\linewidth]{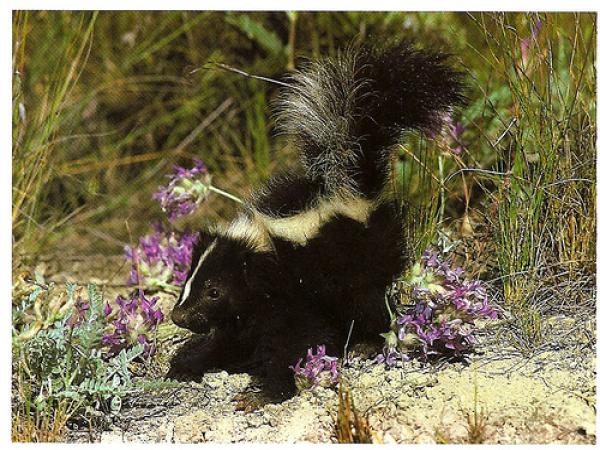}
    }
  \subfloat {
    \includegraphics[width=0.17\linewidth]{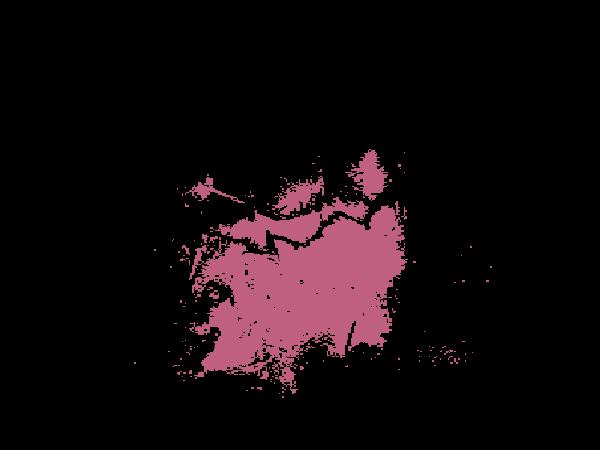}
    }
  \subfloat {
    \includegraphics[width=0.17\linewidth]{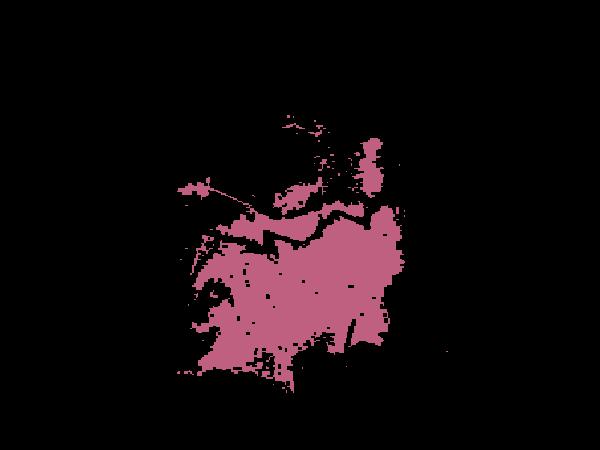}
    }
  \subfloat {
    \includegraphics[width=0.17\linewidth]{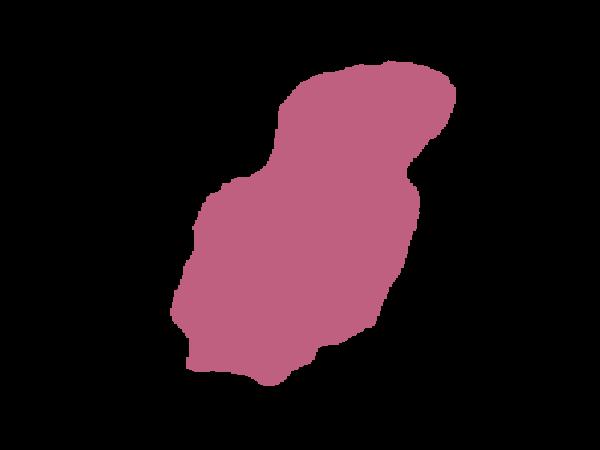}
    }
 \subfloat {
    \includegraphics[width=0.17\linewidth]{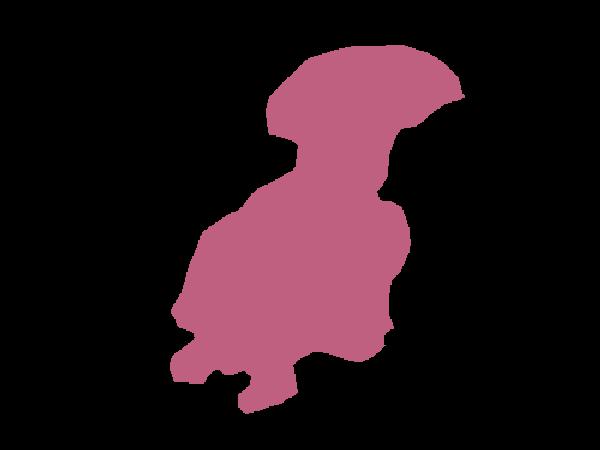}
    }
         \\[-1.5ex]
 \subfloat[(a) Image] {
    \includegraphics[width=0.17\linewidth]{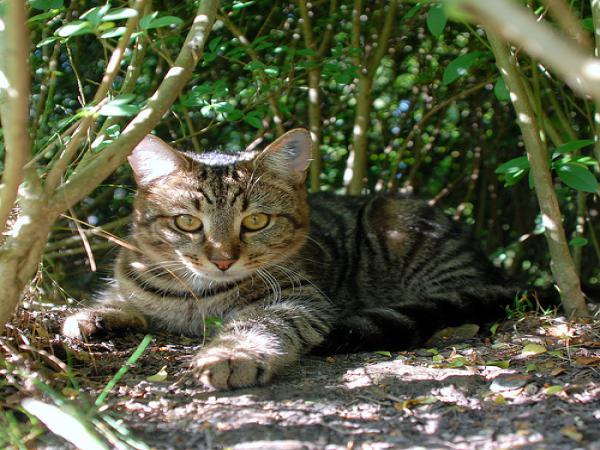}
    }
  \subfloat[(b) Step1. Classification] {
    \includegraphics[width=0.17\linewidth]{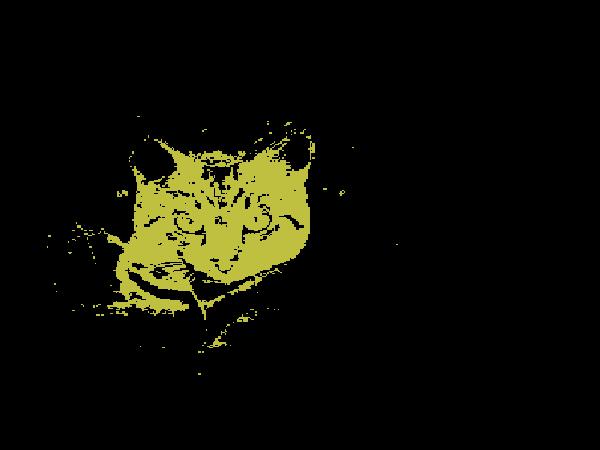}
    }
  \subfloat[(c) Step2. IRNet] {
    \includegraphics[width=0.17\linewidth]{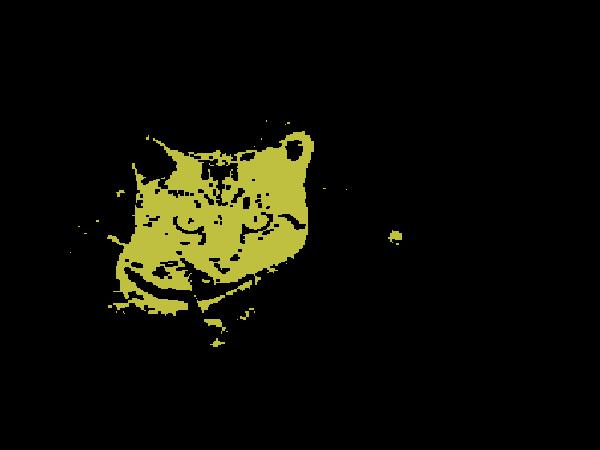}
    }
  \subfloat[(d) Step3. Segmentation] {
    \includegraphics[width=0.17\linewidth]{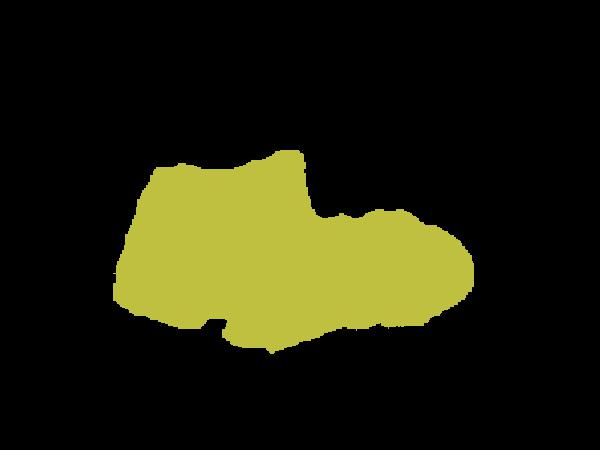}
    }
 \subfloat[(e) Mask] {
    \includegraphics[width=0.17\linewidth]{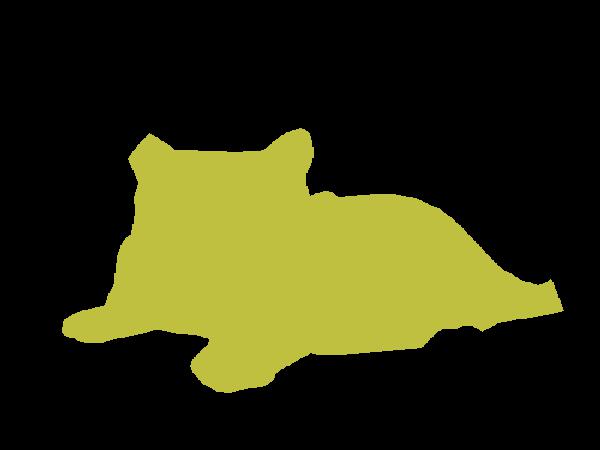}
    }
    \captionof{figure}{Object localization maps for (a) an image at each consecutive step of our method: (b) map after CAM extraction, (c) improved map by IRNet trained on the outcomes of step 1, (d) prediction of DeepLabV3+ trained on step 2 results, all compared to (e) ground truth mask.}
    \label{fig:dataevl}
\end{figure*}

\section{Experiments and Results}

\subsection{Model Evaluation}
For the classification step, we validate our models by calculating the F1 score on image-level labels. This allows us to select the model which performs best in the classification of our extremely imbalanced data. 

The best segmentation model is selected based on the mean IoU achieved on the validation set during the training.

\subsection{Results}
We evaluate performance of our method on the competition server on both validation and test sets using the mean IoU metric. Our method achieves 37.34 mean IoU on the test set, which positions us at the third place. There are two other metrics calculated on the competition server: mean accuracy and mean pixel accuracy. Comparison of top-3 solutions in the LID Challenge is presented in Table \ref{lb}.

\begin{table}[h]
\begin{center}
\begin{tabular}{l|c|c|c}
\hline
Solution & mean IoU & mean accuracy & pixel accuracy \\
\hline\hline
1st & \bfseries 45.18 & 59.62 & 80.46 \\
2nd  & 37.73 & \bfseries 60.15 & 82.98 \\
3rd.Ours & 37.34 & 54.87 & \bfseries  83.64 \\ \hline
\end{tabular}
\end{center}
\caption{Top 3 solutions in the LID Challenge compared using three different metrics.}
\label{lb}
\end{table}

In Table 2, we show how the results improve on validation with each step of our approach

\begin{table}[h]
\begin{center}
\begin{tabular}{l|P{3.4cm}}
\hline
Method Step & mean IoU \\
\hline\hline
Step 1. Classification + CRF & 31.06 \\
Step 2. IRNet & 31.87  \\
Step 3. Segmentation + TTA & 39.64 \\ \hline
\end{tabular}
\end{center}
\vspace{-2.0ex}
\caption{Results on validation at each step of our approach}
\label{stepseval}
\end{table}

We also experiment by testing two encoder architectures of DeepLabv3+  \cite{chen2018encoder} model, different thresholds after IRNet \cite{ahn2019weakly} model, including or excluding 'person' class, and applying various postprocessing techniques. All these experiments are reported in the Table \ref{experimentseval}. 

\begin{table}[h]
\begin{center}
\vspace{1.0ex}
\begin{tabular}{l|c|c|c|c}
\hline
Encoder & IRNet thr. & TTA & Person & mean IoU \\\hline\hline
ResNet50 & 0.3 &  No & No & 36.65 \\ 
ResNet50 & 0.3 &  Yes & No & \bfseries 39.64\\ 
ResNet50 & 0.3 &  Yes & Yes & \bfseries 39.80\\ 
ResNet50 & 0.5 &  No & No & 37.11 \\ 
ResNet50 & 0.5 &  Yes & No & 39.58  \\ 
ResNet101 & 0.5 & No & No & 36.14  \\ 
ResNet101 & 0.5 & Yes & No & 37.15 \\ \hline
\end{tabular}
\end{center}
\vspace{-2.0ex}
\caption{Experiments results on validation set by testing different encoders for DeepLabv3+, two thresholds after IRNet step, using TTA as postprocessing, including CAM for class person from a binary classifier.}
\label{experimentseval}
\end{table}

We provide qualitative results of segmentation on several validation images  in Figure \ref{fig:dataevl}. We show the resulting maps at each step of our method; the figure demonstrates how the performance improves after each step.

\section{Conclusions}
We present a novel method of weakly-supervised semantic segmentation that consists of three consecutive steps: classification, CAM improvement via IRNet, and segmentation. The presented approach generates pseudo-labels from a classifier network, rectifies the class boundaries with IRNet, and uses a supervised segmentation model as a final end-to-end method. This allows us to solve a semantic segmentation task using only image-level annotations.

\subsection{Discussion}

In the proposed approach, the downsampling technique was used to balance the dataset, which was dictated by resource limitations. However, it would be interesting to test upsampling as a class balancing method, or the combination of both. We believe this could give an increase in performance.

Also, we didn't include CAM for class `person' extracted from a binary classifier into the third step - training segmentation model. We think this could be a worthy experiment. 

There is also a space to experiment with different regularization and optimization techniques at all steps. 

\section*{Acknowledgements}
This research was supported by Faculty of Applied Sciences at Ukrainian Catholic University and SoftServe. The authors thank Rostyslav Hryniv for helpful insights, and Tetiana Martyniuk for computational resources.

{\small
\bibliographystyle{abbrv}  
\bibliography{egbib}
}

\end{document}